\pdfoutput=1

\documentclass[11pt]{article}

\usepackage[final]{acl}

\usepackage{times}
\usepackage{latexsym}

\usepackage[T1]{fontenc}

\usepackage[utf8]{inputenc}

\usepackage{microtype}

\usepackage{inconsolata}

\usepackage{graphicx}

\usepackage{hyperref}
\usepackage{url}

\usepackage[utf8]{inputenc} 
\usepackage[T1]{fontenc}    
\usepackage{hyperref}       
\usepackage{url}            
\usepackage{booktabs}       
\usepackage{amsfonts}       
\usepackage{nicefrac}       
\usepackage{microtype}      

\usepackage{pifont}
\usepackage{url}
\usepackage[utf8]{inputenc} 
\usepackage[T1]{fontenc}    
\usepackage{url}            
\usepackage{booktabs}       
\usepackage{amsfonts}       
\usepackage{nicefrac}       
\usepackage{microtype}      
\usepackage{times}
\usepackage{color}
\usepackage{soul}
\usepackage[utf8]{inputenc}
\usepackage{graphicx} 
\usepackage{wrapfig}
\usepackage{amssymb}
\usepackage{amsmath}
\usepackage{multirow}
\usepackage{float}
\usepackage{subcaption}
\usepackage{arydshln}
\usepackage{enumitem}

\usepackage{times}
\usepackage{epsfig}
\usepackage{graphicx}
\usepackage{amsmath}
\usepackage{amssymb}
\usepackage{textcomp}
\usepackage{wrapfig}

\usepackage{bbding}
\usepackage{pifont}
\usepackage{wasysym}
\usepackage{amssymb}
\usepackage{arydshln}
\usepackage{cases}
\usepackage[ruled,vlined,linesnumbered]{algorithm2e}
\usepackage{listings}
\lstalias{plantuml}{}

\title{CodeTree: Agent-guided Tree Search for Code Generation with Large Language Models}

\author{
 \textbf{Jierui Li\textsuperscript{1}}\footnotemark,
 \textbf{Hung Le\textsuperscript{2}},
 \textbf{Yingbo Zhou\textsuperscript{2}},
 \textbf{Caiming Xiong\textsuperscript{2}},
 \textbf{Silvio Savarese\textsuperscript{2}},
 \textbf{Doyen Sahoo\textsuperscript{2}}
\\
 \textsuperscript{1}The University of Texas at Austin
 \\
 \textsuperscript{2}Salesforce Research
\\
 {
\texttt{jierui@cs.utexas.edu}, \texttt{hungle@salesforce.com}
 }}

\begin{document}
\maketitle
\renewcommand{\thefootnote}{\fnsymbol{footnote}}
\footnotetext[1]{The work was done while Jierui Li was a research intern in Salesforce Research Asia.}
\renewcommand{\thefootnote}{\arabic{footnote}}

\begin{abstract}
Pre-trained on massive amounts of code and text data, large language models (LLMs) have demonstrated remarkable achievements in performing code generation tasks. 
With additional execution-based feedback, these models can act as agents with capabilities to self-refine and improve generated code autonomously. 
However, on challenging coding tasks with extremely large search space, current agentic approaches still struggle with multi-stage planning, generating, and debugging.
To address this problem, we propose CodeTree, a framework for LLM agents to efficiently explore the search space in different stages of the code generation process.
Specifically, we adopted a unified tree structure to explicitly explore different coding strategies, generate corresponding coding solutions, and subsequently refine the solutions. 
In each stage, critical decision-making (ranking, termination, expanding) of the exploration process is guided by both the environmental execution-based feedback and LLM-agent-generated feedback. 
We comprehensively evaluated CodeTree on 7 code generation benchmarks and demonstrated the significant performance gains of CodeTree against strong baselines.
Using GPT-4o as the base model, we consistently achieved top results of 95.1\% on HumanEval, 98.7\% on MBPP, and 43.0\% on CodeContests.
On the challenging SWEBench benchmark, our approach led to significant performance gains.

\end{abstract}

\section{Introduction}
Recently, we have witnessed significant impacts of large language models (LLMs) beyond the NLP domain such as in coding tasks \cite{achiam2023gpt, touvron2023llama, wang2023codet5+, roziere2023code}. 
However, different from traditional NLP tasks, coding tasks require generated code to be fully executable and functionally correct i.e. containing no programmatic syntax errors and passing all possible test cases \cite{chen2021evaluating, austin2021program, hendrycksapps2021}. 
Given the extremely large search space in code, early methods propose to sample a very large number of generation outputs (for example, \citet{li2022competition} generated up to 1 million samples per problem) to increase the chance of generating a correct code solution. 
\begin{table*}[t]
\small
\center
\resizebox{1.0\textwidth}{!} {
\begin{tabular}{p{7cm}cccccc}
\hline
\multicolumn{1}{c}{\textbf{Approach}}                 & \textbf{Explore} & \textbf{Exploit} & \textbf{\begin{tabular}[c]{@{}c@{}}Execution \\ feedback\end{tabular}} & \textbf{\begin{tabular}[c]{@{}c@{}}AI \\ feedback\end{tabular}} & \textbf{\begin{tabular}[c]{@{}c@{}}Multi-\\ agent\end{tabular}} & \textbf{Action}\\
\hline
CodeRanker \cite{inala2022faultaware}                                           & \checkmark              &                  &                                                                        & \checkmark                                                             &                                                                 &\\
AlphaCode \cite{li2022competition}, MBR-Exec \cite{shi-etal-2022-natural}, CodeT \cite{chen2023codet}             & \checkmark              &                  & \checkmark                                                                    &                                                                 &                                                              &   \\
LEVER \cite{ni2023lever}, Coder-Reviewer \cite{zhang2023coder}                                 & \checkmark              &                  & \checkmark                                                                    & \checkmark                                                             &                                                             &    \\
\hline
Self-correct \cite{welleck2023generating}, ILF \cite{chen2023improving}, Self-refine \cite{madaan2023self}                       &                  & \checkmark              &                                                                        & \checkmark                                                             & &                                                                \\
CodeChain \cite{le2024codechain}                                            &                  & \checkmark              & \checkmark                                                                    &                                                                 &                                                             &    \\
Self-debug \cite{chen2023teaching}, Self-repair \cite{olausson2023demystifying}, Reflexion \cite{shinn2023reflexion} &                  & \checkmark              & \checkmark                                                                    & \checkmark                                                             &                                                             &    \\
\hline
CAMEL \cite{li2023camel}                                                &                  &                  &                                                                        & \checkmark                                                             & \checkmark                                                             \\
ChatDev \cite{qian2024chatdev}, MetaGPT \cite{hong2023metagpt}, AgentVerse \cite{chen2023agentverse}                         &                  &                  & \checkmark                                                                    & \checkmark                                                             & \checkmark                                                             \\
Self-collaboration \cite{dong2023self}, AgentCoder \cite{huang2023agentcoder}                       &                  & \checkmark              & \checkmark                                                                    & \checkmark                                                             & \checkmark       &                                                      \\
\hline
CodeTree (ours)                                             & \checkmark              & \checkmark              & \checkmark                                                                    & \checkmark                                                             & \checkmark &\checkmark \\
\hline
\end{tabular}
}
\caption{
We compare CodeTree with related methods in 6 aspects: 
(1) \emph{Explore} which adopts a brute-force approach to independently generate a large number of code candidates;
(2) \emph{Exploit} which focuses on self-refinement using a small subset of output solutions;
(3) \emph{Execution feedback} which uses code execution outcomes to improve code qualities;
(4) \emph{AI feedback} which enables synthetic feedback generated by LLMs to improve output code;
(5) \emph{Multi-agent} which adopts multiple LLM agents to play different roles in the code generation process; and
(6) \emph{Action} where LLM agents can take different actions and facilitate decision-making.
} 
\label{tab:related_work}
\end{table*}

\begin{figure*}[htbp]
  \includegraphics[width=\textwidth]{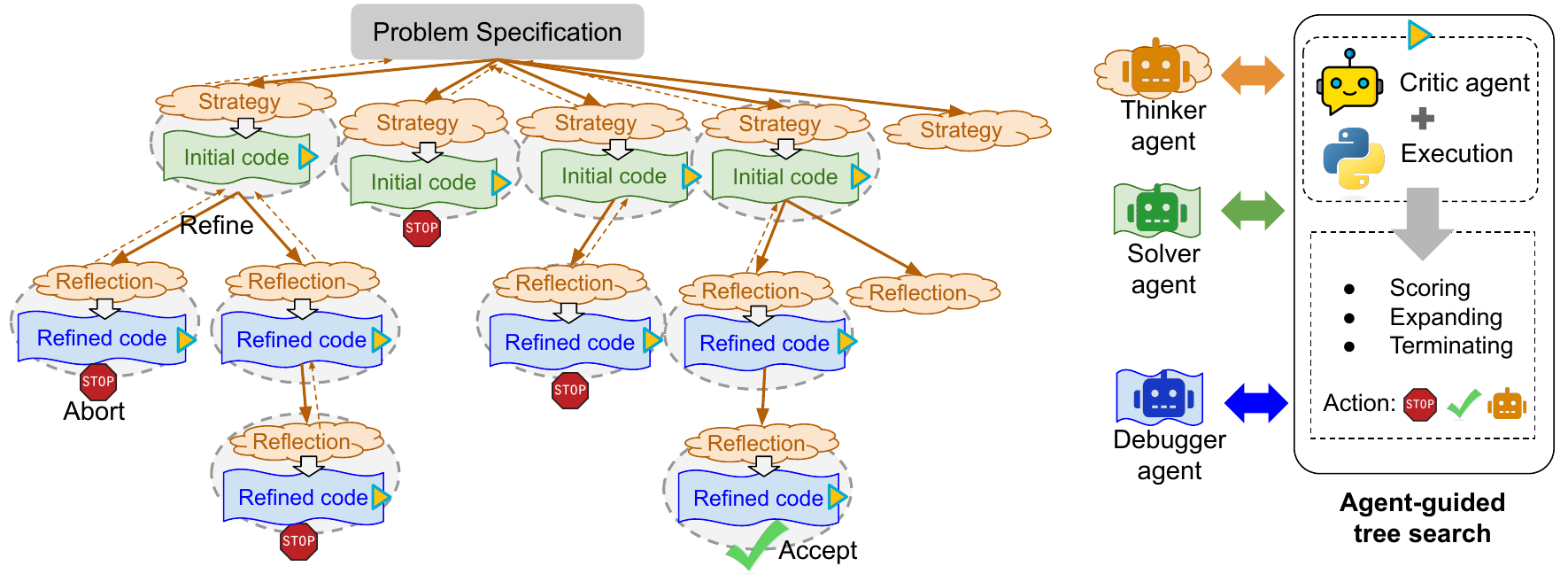}
  \caption{
   CodeTree creates a unified search space for exploration throughout the multi-stage code generation process: strategy generation by a ``Thinker'' agent, initial code generation by a ``Solver'' agent, and code improvement by a ``Debugger'' agent. 
   To effectively perform exploration within the tree structure, we incorporate both environmental execution-based feedback as well as AI-generated feedback (generated by a ``Critic'' LLM agent).
  }
  \label{fig:method}
\end{figure*}

More recently, several approaches adopted a ``vertical'' strategy in which LLMs first generate one (or very few) generation output, and then iteratively refine this output multiple times, often conditioned by some forms of external feedback \cite{le2022coderl, chen2023codet, shinn2023reflexion}. 
While these approaches are more cost-effective by focusing only on a small subset of the search space (i.e. starting from an initial output candidate), the performances of these approaches are bounded by the local optima of the chosen search space. 
Related to our work, several methods for NLP reasoning tasks were introduced to control and enhance the generation procedure of LLMs. 
For example, \citet{wang2022self} proposed to enhance LLMs with chains of thought and statistically select the right solutions based on majority voting. 
\citet{zhou2023leasttomost} decomposed a task into smaller sub-tasks and addressed them by increasing the order of difficulty. 
\citet{yao2024tree} proposed to improve LLMs by adopting a tree-based structure to explicitly simulate the exploration of thoughts in a tree.
We are motivated by this line of research and proposed CodeTree, a new generation framework to effectively explore the search space of code generation tasks through a tree-based structure. 
An overview of CodeTree is given in Figure \ref{fig:method}.

We define 3 standard agents, Thinker, Solver, and Debugger, to equip the strategy-planning, solution implementation, and solution improvement correspondingly, posing comprehensive roles needed for code generation. A CodeTree starts from the input problem as the tree root and subsequent nodes represent code solutions. At any node of the tree, one can either explore sibling nodes (other strategies from the same parent node) or its children (refinements of this node). Within CodeTree,  agents can interact with each other through a tree expansion guided by a Critic Agent, searching for the optimal code solution. 

Rather than following heuristic rules or classic tree traversal methods, we use Critic Agent to self-evaluate the status of tree nodes at each tree expansion step by executing the following tasks: 
\begin{itemize}
    \item Node scoring: Evaluate the test case outputs of generated code and assess whether the tree nodes faithfully follow their corresponding coding strategies.
    \item Solution verification \& evaluation: For a solution that passes visible test cases, verify if it should be further improved; for a solution that fails, evaluate if it is a promising direction to debug and refine the solution. 
\end{itemize}
Based on the outputs of the above evaluation tasks, Critic Agent can take action to either refine, abort, or accept the current solution, which automatically expands or terminates the tree search. CodeTree is flexible and efficient, avoiding duplicated or redundant exploration of functionally similar or unfeasible solutions.

We comprehensively evaluated CodeTree on diverse code generation benchmarks from beginner- to competition-level coding tasks. 
Our results demonstrated the significant and consistent performance gains of CodeTree against strong baselines.
Using GPT-4o as the base models, we achieved the top results on HumanEval+, MBPP+ \cite{evalplus}, and CodeContests \cite{li2022competition} respectively. 
On the challenging SWEBench benchmark, our approach led to significant performance gains. 
We also conducted comprehensive ablation and qualitative analysis to derive the best practices and any limitations of the current method. 

\section{Related Work}

Our work is broadly related to the research of large language models (LLMs) \citep{palm, achiam2023gpt, touvron2023llama}.
\citet{roziere2023code, li2023starcoder, wang2023codet5+, chen2021evaluating} extended this line of research by allowing LLMs to learn from large-scale code-related data such as open-sourced Github repositories and code commits. 
Treating code generation as an autoregressive generation tasks, LLMs can generate code that correctly follow programming syntactic rules and are functionally correct \cite{chen2021evaluating, gunasekar2023textbooks, nijkamp2023codegen2}. 
Early approaches \citep{chen2021evaluating, austin2021program, hendrycksapps2021} adopted a brute-force principle by independently generating a very large number of output code candidates, so one of which might be the optimal code solution.
Subsequently, \citet{li2022competition, chen2023codet, ni2023lever} proposed to  utilize unit test outcomes to filter for more prospective generated code samples. 

Also related to our work is the studies of self-refinement capabilities of LLMs.
These studies leverage the inherent ability of LLM to perceive arbitrary natural language contexts, including information such as environmental feedback, to iteratively fix problematic code and improve its correctness.
For example, \citet{zhang2023self} utilized test outcomes as a form of feedback while \citet{welleck2023generating, le2022coderl} introduced LLM-generated feedback from predicted probabilities about code correctness. 
\citet{shinn2023reflexion, chen2023teaching, madaan2023self} focused on more natural forms of feedback such as reflections and explanations of the code.
\citet{le2024codechain} proposed to cluster sub-module representations of code as a form of collective feedback. 

More related to our work is the research for enhancing and controlling the generation procedure of LLMs.
\citet{yao2024tree, koh2024tree} simulated the step-by-step exploration of thoughts as a tree search to support NLP and multimodal tasks respectively. 
In the code domain, \citet{song2024effective} used tree search to guide the debugging of generated code.  
\citet{islam2024mapcoder} adopted multi-agent system to support different stages of code generation but the exploration is not explicitly defined. 
\citet{chen2024divide} introduced a tree structure to explore sub-functions in generated code. 
Different from prior approaches, we introduce a tree-based structure as a unified search space for LLM agents to efficiently perform exploration throughout different stages of code generation. 
See Table \ref{tab:related_work} for a more systematic comparison with related methods. 

\section{Method}

To apply LLMs to code generation, we can define this task as a sequence-to-sequence task. The input sequence consists of a problem description $D$, usually in the form of a function docstring (including expected input and output) or the textual explanation of the problem. 
The output is a corresponding code solution, flattened into a sequence of tokens $\hat{W}=(\hat{w}_1, ...,\hat{w}_T)$ with $\hat{w}_t \in \mathcal{V}$.

Generated codes are evaluated against hidden test cases to check the functional correctness \citep{hendrycksapps2021, chen2021evaluating, li2022competition}. 
The test cases are a set of input-output pairs $\{(i_j, o_j)\} = \{ (i_j, o_j)_v\} \cup \{(i_j, o_j)_h\} $. Visible test cases are denoted as  $\{(i_j, o_j)_v\}$ while hidden test cases are denoted as $\{(i_j, o_j)_h\}$. An output code $\hat{W}$ is correct when $\hat{W}(i_j)=o_j$ $\forall j$.

\begin{figure}[ht]
    \centering
        \includegraphics[width=\columnwidth]{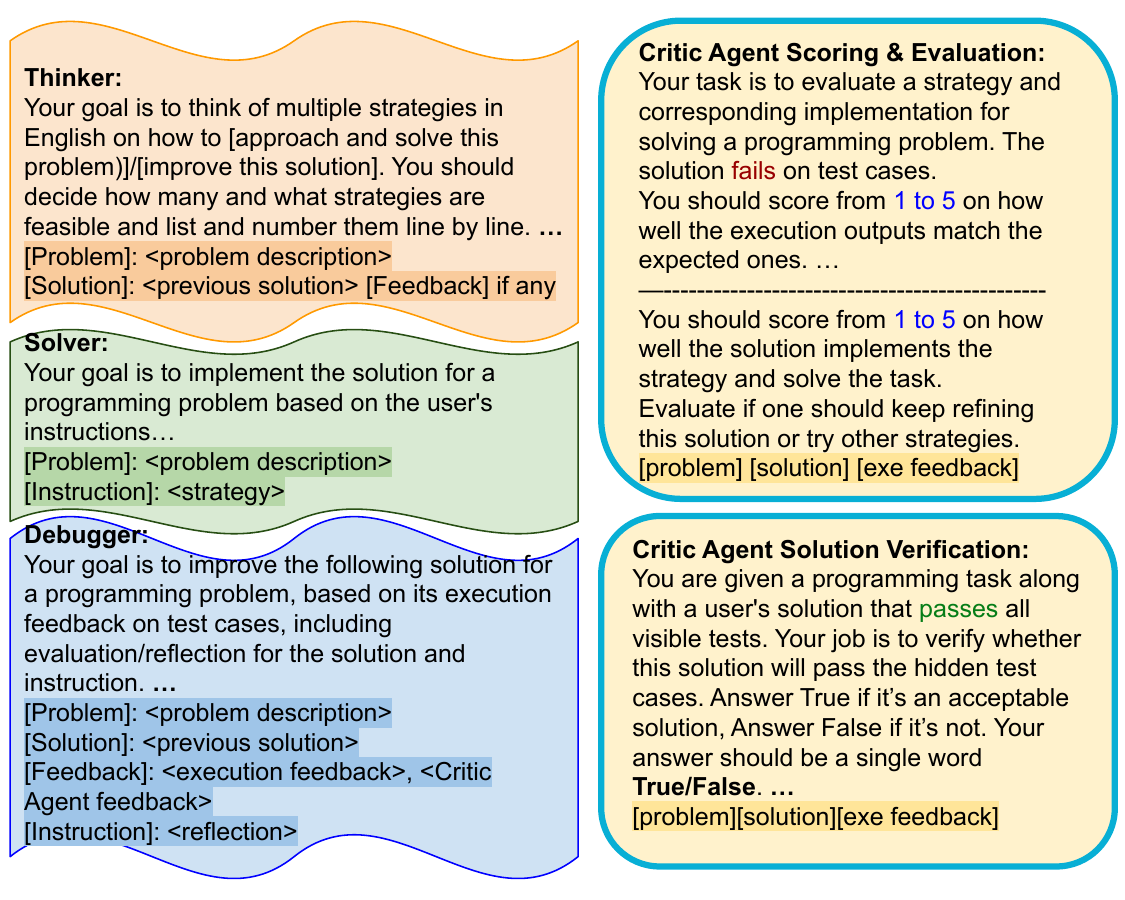}
        \caption{Simplified versions of instruction prompts used for Thinker, Solver, Debugger, and Critic agents. Some details are omited for illustration purposes.}
        \label{fig:prompt}
\end{figure}

Please refer to Figure \ref{fig:method} for an overview of our method and Figure \ref{fig:prompt} for a simplified version of instruction prompts to our LLM agents. 

\subsection{Coding Task-Specific Agents}
We first introduce three unit LLM agents, specifically targeting different parts of the code generation process, including strategy thinking, code implementation, and code debugging. 

\paragraph{Strategy Generation with Thinker Agent}
\label{subsec:thinker}
Conventional approaches such as \cite{chen2021evaluating, austin2021program, hendrycksapps2021} directly generate output code candidates given a problem description. 
However, these approaches do not fully exploit the advantage of LLMs in generating more expressive outputs in text. 
\citet{wei2022chain} showed that allowing LLMs to generate natural language step-by-step thoughts can lead to significant performance gains in downstream NLP tasks. Following the setting in \citet{yao2024tree}, requesting LLMs to generate a list of different natural language \textit{thoughts} can enhance the diversity of solutions. 
We propose to adapt this technique to allow an LLM $\theta_{T}$ (denoted as ``Thinker'' agent) to sequentially generate a set of high-level strategies given an input coding problem. 
Each strategy $\hat{S_i}$ is generated autoregressively given previously generated strategies following: 
\begin{align}
    \hat{S_i} &\sim p_{\theta_{T}} (.| \hat{S}_{1:i-1}, D) \label{eq:thinker}
\end{align}
By allowing models to first generate coding strategies, we enable LLMs to tackle coding problems using their reasoning capabilities learned from the text domain.
The expressiveness of generated strategies in a natural language can potentially guide the code-generation process toward more diverse exploration paths. 
Notably, we let Thinker Agent dynamically decide the number of generated coding strategies, given the fact that different coding problems can have more or fewer feasible strategies. 

\paragraph{Solution Generation with Solver Agent}
\label{subsec:solver}
Given a complete generated strategy $\hat{S_i}$, we let an LLM $\theta_{S}$ (denoted as ``Solver'' agent) generate a set of initial code solutions.
Since LLMs are often fine-tuned to follow arbitrary instructions in natural language, these models can understand novel unseen tasks during test time \citep{ouyang2022training, touvron2023llama2, wang2023codet5+}.  
Therefore, by including the strategy as part of the input instruction, we can condition Solver Agent to produce strategy-specific code candidates. For each candidate, the generation objective is defined as:
\begin{align}
    \hat{W_i} &\sim p_{\theta_{S}} (\hat{S_i}, D)  \label{eq:solver}
\end{align}

\paragraph{Solution Refining with Thinker \& Debugger Agents}
\label{subsec:debugger}
Prior approaches such as \citep{chen2023improving, chen2023teaching, shinn2023reflexion, madaan2023self} found that syntactic mistakes or even logical flaws in generated code can be fixed by allowing LLMs to iteratively refine and regenerate the code. 
This self-refinement capability is typically strengthened by some forms of feedback about the code qualities (e.g. execution results, compiler signals):
\begin{align}
    F_{exe,i} &= \hat{W_i}(\{(i_j, o_j)_v\}) \\
    F_{cri,i} &= {\theta_{C}}(\hat{W_i}, \hat{S_i} , F_{exe, i}, D) \label{eq:feedback}
\end{align} 
where ${\theta_{C}}$ is Critic Agent. Denoting the collective feedback as $F_i=\{F_{exe,i}, F_{cri,i}\}$,  a set of reflections $R_i$ about the code candidates are generated by Thinker Agent. 
\begin{align}
    \hat{R}_{i,j} &\sim p_{\theta_{T}} (.| \hat{R}_{i,1:j-1}, F_i, \hat{W}_i, \hat{S_i}, D) \label{eq:reflection}\\
    \hat{W}_{i,j} &\sim p_{\theta_{D}} (.| \hat{R}_{i,j}, F_i, \hat{W}_i, \hat{S_i}, D) \label{eq:debugger}
\end{align}
$\hat{R}_{i,j}$ denotes the $j$th reflection that ``Thinker'' generates for $\hat{W_i}$. An LLM $\theta_{D}$ (``Debugger'' Agent) will modify $\hat{W}_i$, referring this reflection, generating a new program correspondingly. 

\subsection{Tree Expanding with Critic Agent}
\label{sec:critic}
CodeTree builds a heterogeneous tree for each problem, where the tree root represents a problem specification $(D, \{(i_j, o_j)\})$ and every subsequent tree node represents a generated code solution $\hat{W}_i$.
Each node has relevant attributes including its collective code feedback $F_i$ and its corresponding strategy and reflections: $S_i$ and $R_i$. Typically, adding a tree node is a two-step process: 1) generate a code solution from the corresponding strategy (Eq. \ref{eq:solver} or Eq. \ref{eq:debugger}), 2) evaluate the generated solution $\hat{W}_i$ and obtain environmental feedback (Eq. \ref{eq:feedback}). 

Unlike previous tree-structure search methods \cite{yao2024tree, islam2024mapcoder}, we do not construct the entire tree in the beginning. Instead, we introduce a Critic Agent to dynamically expand the tree based on potential strategies. It will guide the expansion and spanning of the tree, taking actions based on its evaluation of the current node. 

\paragraph{Node Scoring and Evaluation}
For a given solution and corresponding $F_{exe}$, Critic Agent performs an evaluation, measuring how promising it is through equation \ref{eq:feedback}, which results in $F_{cri}$. 
We separately evaluate how well: 1)  the execution outputs of test cases match expected outputs on visible test cases; and 
2) the solution robustly implements its corresponding strategy towards problem-solving.
For one program $\hat{W}_i$ and its corresponding feedback $F_i$, the Critic Agent will evaluate whether the current solution is worth refining, or it should not be explored, making decision between refinement and abort.
The critic score is calculated following the equation:  
\begin{align}
    \text{Score}(\hat{W}_i) = \text{Score}(F_{exe,i}) + \text{Score}(F_{cri,i})
\end{align}

\paragraph{Solution Verification}
For one $\hat{W}_i$ that passes all visible test cases, it might potentially over-fit the visible test cases and could fail hidden test cases. Hence, the critic agent ${\theta_{C}}$ will verify if this solution is robust and generalizable to unseen test cases.

\paragraph{Decision-Making by Critic Agent}
Starting from the initial $S_i, W_i, F_i$, Critic Agent guides the search for a correct solution. 
At each node, it has an action space of three actions:
\textbf{Refine}: Continue exploring from the current node by generating multiple reflections for this node;
\textbf{Abort}: Prune this node due to its low evaluation score, and retrace the exploration to its sibling nodes; and 
\textbf{Accept}: Accept the current node as the final solution and terminate the search early.


\subsection{Multi-agent Collaboration}
Throughout the expansion of the tree, the task-specific agents collaborate with Critic Agent, utilizing its feedback, and follow its guidance to perform exploration. 
The flexibility of the tree expansion and search is determined by LLM agents' decision-making, e.g. determining the number of strategies and deciding the search path. 
During inference time, practically, we limit the number of exploration steps to avoid large computation overhead. 
Whenever a termination signal (i.e. to accept a code solution) is found or the maximum number of exploration steps is reached, a code candidate is selected based on its evaluation score $\text{Score}\hat{(W_i)}$. 
Please refer to Appendix \ref{sec:appendix} for all example instruction prompts of our LLM agents.

\section{Experiments}
\begin{table*}[ht]
  \centering
  \small
    \begin{tabular}{llcccccc}
        \toprule
        \textbf{Model} & \textbf{Method} & \textbf{HumanEval} & \textbf{HumanE+} & \textbf{MBPP} & \textbf{MBPP+} & \textbf{Codecontests} &\textbf{Avg.} \\
        \midrule
        \multirow{9}{*}{\textbf{GPT-4o-mini}} 
            & \cellcolor{lightgrey}Direct & 86.6\% & 78.7\% & 87.8\% & 73.3\% & 10.3\% &67.3\%\\
            & \cellcolor{lightgrey}CoT  & 84.8\% & 78.0\% & 89.2\% & 74.3\% &12.7\% & 67.8\% \\
            \cmidrule(r){2-8} 
            & \cellcolor{lightgreen}Reflexion & 92.1\%& 83.5\% & 96.6\% & \textbf{78.6\%} & 21.8\% &74.5\% \\
            & \cellcolor{lightgreen}MapCoder  & 91.5\% & 78.0\%& 90.0\% & - & - & -\\
            & \cellcolor{lightgreen}Resample      & 89.0\%      & 80.5\%    & 94.3\% & 76.8\%    & 18.2\% &71.8\% \\
            \cmidrule(r){2-8}
            &\cellcolor{lightyellow}CodeTree-BFS           & 93.3\%    & 82.1\%    & 91.5\%           & 72.3\%    & 20.6\% & 72.0\% \\
            & \cellcolor{lightyellow}CodeTree-DFS           & 92.7\%    & 81.1\%    & 87.6\%           & 71.4\%    & 20.6\% & 70.7\%\\
            & \cellcolor{lightyellow}Strategy List & 90.2\%    & 80.5\%    & 90.5\%           & 69.6\%    & 22.4\% & 70.6\% \\
            & \cellcolor{lightyellow}CodeTree          & \textbf{94.5\%} & \textbf{84.8\%} & \textbf{96.8\%}           & 77.0\% & \textbf{26.4\%} & \textbf{75.9\%} \\
        \midrule
        \multirow{9}{*}{\textbf{GPT-4o}}
            & \cellcolor{lightgrey}Direct & 88.4\% & 81.7\% & 92.3\% & 75.9\% & 20.6\% & 71.8\%\\
            & \cellcolor{lightgrey}CoT  & 92.1\%& 84.1\% & 93.7\% & 77.2\% & 24.8\% & 74.4\% \\
            \cmidrule(r){2-8}
            & \cellcolor{lightgreen}Reflexion & 94.5\% & 84.8\% & 97.9\% & 79.6\% & 41.8\% &79.7\% \\
            & \cellcolor{lightgreen}MapCoder  & 92.7\%& 81.7\% &90.9\% &- & - & -\\
            & \cellcolor{lightgreen}Resample      & 93.9\% & 84.8\%   & 96.2\%           & 77.0\%      & 32.7\% & 76.9\%\\
            \cmidrule(r){2-8}
            & \cellcolor{lightyellow}CodeTree-BFS           & 94.5\%    & 84.1\%    & 93.9\%           & 70.7\%      & 35.8\% & 75.8\% \\
            & \cellcolor{lightyellow}CodeTree-DFS           & \textbf{95.1\%}    & 83.5\%    & 91.5\%           & 76.2\%    & 36.4\% & 76.5\% \\
            & \cellcolor{lightyellow}Strategy List & \textbf{95.1\%}   & 82.3\%    & 92.6\%           & 73.3\%    & 36.4\% & 75.9\%\\
            & \cellcolor{lightyellow}CodeTree          & 94.5\% & \textbf{86.0\%}    & \textbf{98.7\%} & \textbf{80.7\%} & \textbf{43.0\%} & \textbf{80.6\%} \\
            \midrule
            \multirow{9}{*}{\textbf{Llama-3.1-8B}}
            & \cellcolor{lightgrey}Direct & 63.4\% & 54.3\% & 73.4\%& 63.8\% & 6.1\% & 52.2\%\\
            & \cellcolor{lightgrey}CoT  &65.9\% & 56.1\% & 74.6\% & 65.3\% &4.2\% & 53.2\%\\
            \cmidrule(r){2-8}
            & \cellcolor{lightgreen}Reflexion &79.9\%  & 69.5\%& 90.2\%& 72.0\% & 13.5\% & 65.0\% \\
            & \cellcolor{lightgreen}Resample      & \textbf{82.3\%} & 71.3\%   & \textbf{91.0\%}           & \textbf{73.8\%}      & \textbf{15.2\%} &\textbf{66.7\%}\\
            \cmidrule(r){2-8}
            & \cellcolor{lightyellow}CodeTree-BFS           & 80.5\%    & 68.3\%    & 91.0\%           & 69.3\%      & 15.8\% &65.0\% \\
            & \cellcolor{lightyellow}CodeTree-DFS           & 80.5\%    & 68.9\%    & 89.7\%           & 70.4\%    & 15.2\% & 64.9\%\\
            & \cellcolor{lightyellow}Strategy List & 82.3\%    & 70.1\%    & \textbf{91.0\%}          & 72.5\%    & 13.9\% &66.0\%\\
            & \cellcolor{lightyellow}CodeTree  &82.3\% & \textbf{72.0\% }   &  90.5\% &  73.3\% & 12.1\% &66.0\%\\
        \bottomrule
    \end{tabular}
    \caption{\label{tab:main_res}
    Experimental results by pass@1 on HumanEval, MBPP, EvalPlus, and CodeContests: \colorbox{deepgrey}{~~} methods are baseline methods that generate program solution only once, \colorbox{deepgreen}{~~} are methods with solution generation budget of 20 samples like our methods.
    \colorbox{deepyellow}{~~} are CodeTree variants with or without Critic Agent to guide the tree search.  Note that 
MapCoder does not work with Llama-3.1-8B as noted by \citet{islam2024mapcoder}.}
\end{table*}
\subsection{Experimental Setup}
We applied pass@1\cite{chen2021evaluating} as our evaluation metric: only one code candidate can be selected and submitted for the final evaluation with hidden test cases. 
We set the generation budget to be $20$ samples per coding task. 
To fairly compare our approach with other baselines, we adopted the same generation budget in all methods. 
For ablation experiments without using Critic Agent, we followed similar strategies from \cite{shinn2023reflexion, chen2023teaching}: we select a solution which passes all visible test cases as the final solution to be evaluated with hidden test cases. 

\paragraph{Benchmarks} We conducted experiments on 2 categories of code generation tasks: 1) Function implementation where a coding task is to complete a single function following a specific function signature: HumanEval \cite{chen2021evaluating}, MBPP \cite{austin2021program}, and their EvalPlus variants from \cite{evalplus}, denoted as HumanEval+ and MBPP+; and 2) Program implementation where a coding task is to solve an algorithmic problem:  CodeContests \cite{li2022competition} and APPS \cite{hendrycksapps2021}. The sizes of test set are 164, 378 and 165 for HumanEval(+), MBPP(+) and CodeContests respectively. For APPS, we randomly sampled 150 samples (50 for each level of difficulty) from the test split. 

\paragraph{Baselines} We introduce the following baselines: 
\textbf{Direct} instructs the model to generate code directly from the input problem; 
\textbf{CoT} \cite{wei2022chain} instructs the model to provide chain-of-thought reasoning before giving the solution program;
\textbf{Reflexion} \cite{shinn2023reflexion} utilizes solution's execution feedback to generate self-reflections. The reflections are used to iteratively refine the solution;
\textbf{MapCoder} \cite{islam2024mapcoder} proposes an agent collaboration system to plan, solve, test, and refine the solution. We set \#plans=4, \#debug-round=5 and generation budget=20; and
\textbf{Resample} follows a similar principle as \citet{li2022competition}: resample solutions repeatedly and filter them with visible test cases.\footnote{We set sampling temperature=1 for Resample, and report the best results over 2 runs. For other methods, we report the single run's results with deterministic inference.}

\paragraph{Models} We studied our method on three models with different model sizes and capacities. We experimented on large language models from the GPT and Llama 3.1 family.
Specifically we use \textbf{GPT-4o-mini}, \textbf{GPT-4o}\footnote{\url{https://openai.com/index/hello-gpt-4o/}},
and 
\textbf{Llama-3.1-8B }\footnote{\url{https://huggingface.co/meta-llama/Llama-3.1-8B-Instruct}. Note that we reported our replicated results which might be different from the original reported ones.}.

\subsection{Main Results}

We compared CodeTree with other baselines in Table \ref{tab:main_res}. We noticed that Reflexion and Resampling serve as strong baselines for HumanEval and MBPP datasets given the same solution generation budget, comparable to CodeTree-BFS/DFS. CodeTree with Critic Agent outperforms all other baselines in 4 out of 5 benchmarks for GPT-4o-mini and GPT-4o. For instance, CodeTree achieves pass@1=43.0\% on competition-level coding tasks in the Codecontests benchmark (i.e. 22.4\% performance gain over the Resampling baseline), showing its advantage in solving hard problems. 

We found that CodeTree-BFS almost always performs better than CodeTree-DFS, suggesting that exploring diverse strategies is more effective than iteratively refining from one solution. 
Interestingly, on Llama-3.1-8B model, Resampling achieves the best results on 4 benchmarks. 
This observation indicates that small language models may not be suitable for multi-agent frameworks like CodeTree, where models are required to follow task-specific roles and instructions and perform distinct tasks with reasonable accuracy. 

\subsection{Analysis of Search Strategies}
\begin{table}[htbp]
    \centering
    \small
    \resizebox{1.0\columnwidth}{!} {
    \begin{tabular}{lccc}
        \toprule
        \textbf{Model} & \multicolumn{2}{c}{\textbf{GPT-4o-mini}} & \textbf{GPT-4o}  \\
        \cmidrule(lr){2-4}
        \textbf{Benchmark} & \small HumanEval & \small HumanEval+ & \small CodeContests \\
        \midrule
        \multicolumn{2}{l}{\textbf{CodeTree-BFS}} & &  \\
        $d=3,w=3$& 93.3\% & 82.3\% & 36.4\% \\
        $d=2,w=4$& \textbf{95.1\%} & \textbf{84.1\%} & 37.6\% \\
        $d=2,w=5$& 94.5\% & 83.4\% & \textbf{39.4\%} \\
        \midrule
        \multicolumn{2}{l}{\textbf{CodeTree-DFS}} & &  \\
        $d=3,w=3$& 92.7\% & 81.1\% & 36.4\% \\
        $d=4,w=2$& 92.1\% & 81.1\% & 37.0\% \\
        $d=5,w=2$& 92.1\% & 81.7\% & 36.4\% \\
        \bottomrule
    \end{tabular}
    }
    \caption{Pass@1 results of CodeTree-BFS/DFS on HumanEval, HumanEval+, and CodeContests: $d$ indicates max search depth while $b$ indicates max search width. }
    \label{tab:width}
\end{table}

\begin{figure*}[htbp]
    \centering
        \includegraphics[width=0.95\linewidth]{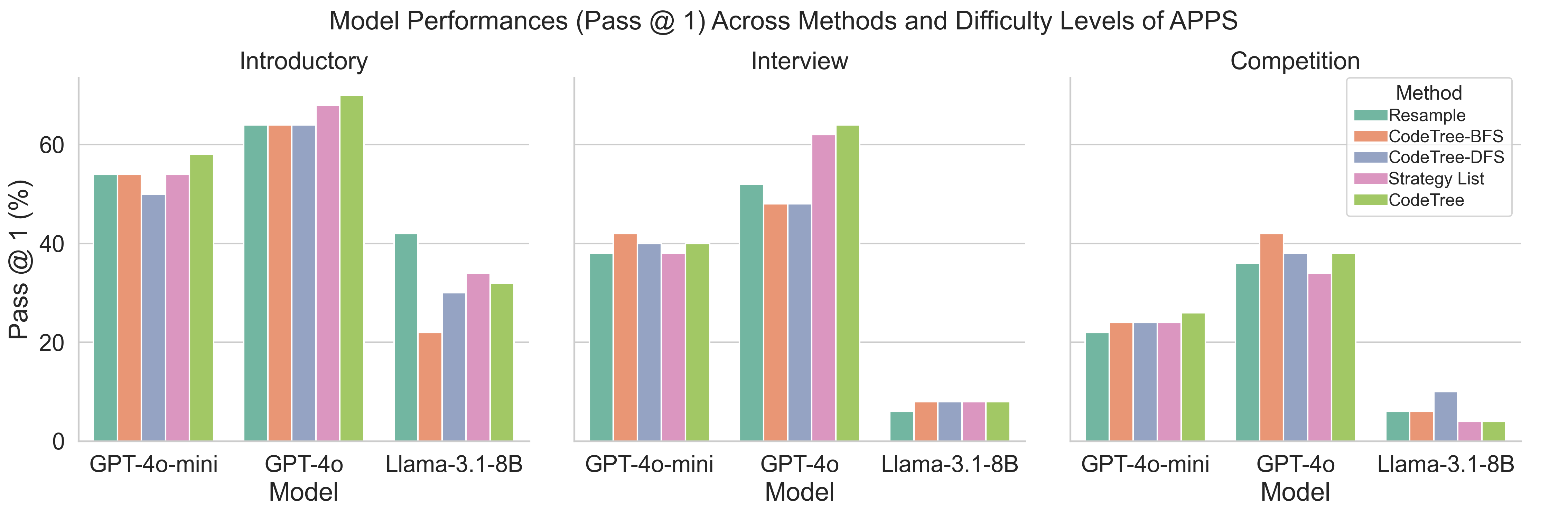}
        \caption{Results of pass@1 on the APPS test subsets. 
        We randomly select 50 samples from Introductory, Interview, and Competition separately.
        We apply Resample and our methods with GPT-4o, GPT-4o-mini, Llama-3.1-8B.}
        \label{fig:apps}
\end{figure*}
Given the performance gaps between CodeTree-BFS and DFS, we conducted additional experiments to analyze these tree search strategies without Critic Agent. We reported the results on HumanEval/HumanEval+ with GPT-4o-mini and Codecontests with GPT-4o in Table \ref{tab:width}. 
Compared to DFS/BFS strategies with $d=3$ and $w=3$, we observed that forcing the model to search wider (i.e. more diverse strategies with $w>3$) in BFS and only debug up to 1 iteration (i.e. $d=2$) improved the performance of pass@1.
However, for DFS, prioritizing deeper search (i.e. larger number of iterations for refinement with $d>3$) does not boost the performance significantly. 
Finally, we noted that $w=4$ works better for HumanEval and $w=5$ works better for CodeContests, indicating that more complex problems can benefit from exploring a larger number of coding strategies. 
This finding supports our proposed CodeTree in which Critic Agent can dynamically determine the number of child nodes to explore given the context.
\subsection{Analysis by Problem Difficulty}
We further evaluated CodeTree against coding problems with diverse difficulty levels. We randomly sampled 50 problems from each level of difficulty (``introductory'', ``interview'', and ``competition'') from the test split of the APPS benchmark, creating a total test set of 150 problems. We reported the results in Figure \ref{fig:apps}.

The results demonstrate that CodeTree performs better for simpler problems (Introductory for GPT-4o-mini; Introductory and Interview for GPT-4o), but still struggles to solve very hard problems (e.g., Competition-level). 
We hypothesized that while CodeTree improves the search efficiency towards the correct solution, a generation budget$=20$ is still limited for highly complex problems. 

\begin{figure*}[ht]
    \centering
    \begin{subcaptionblock}{0.49\textwidth}
        \includegraphics[width=\linewidth]{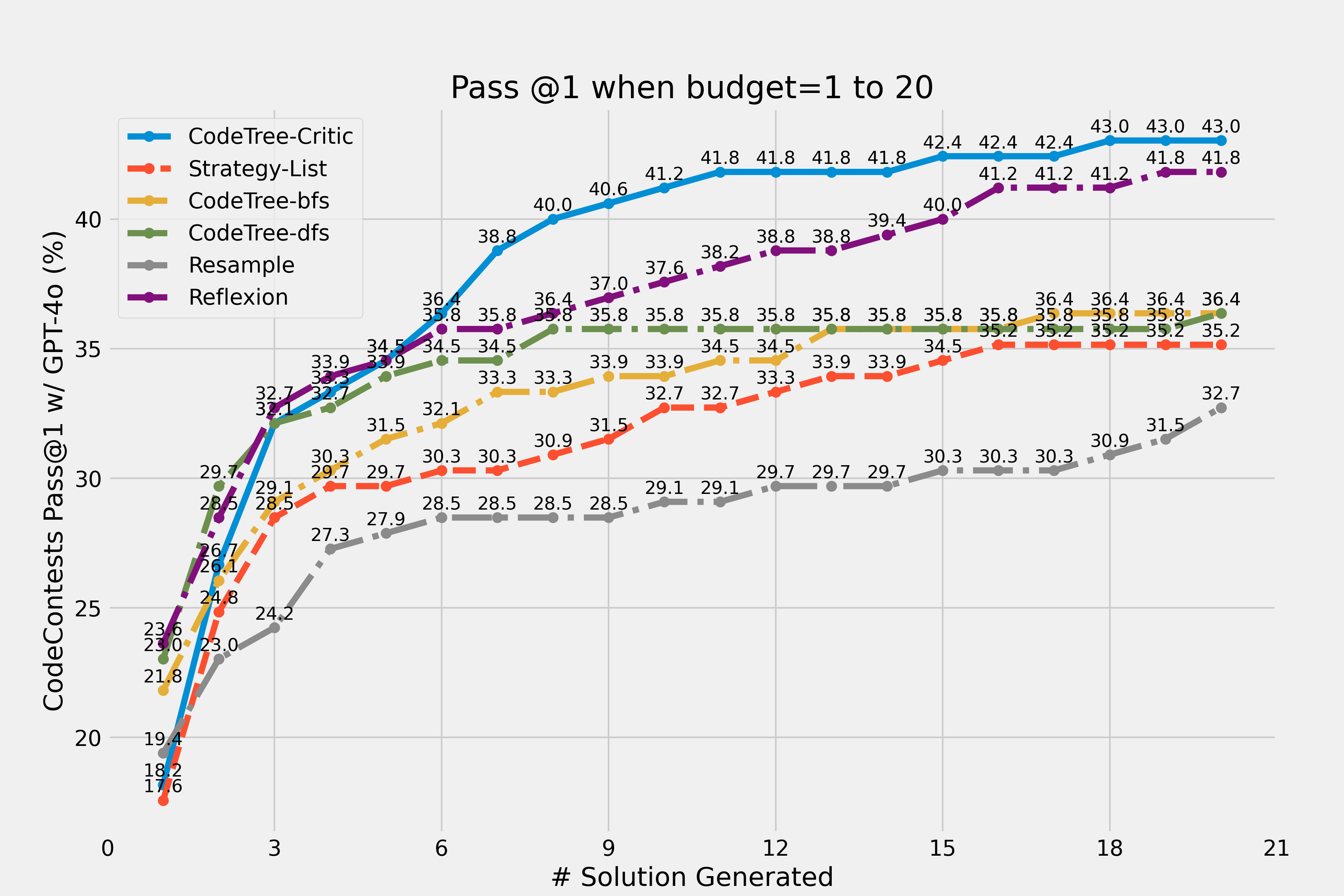}
        \caption{Performance on CodeContests with GPT-4o}
    \end{subcaptionblock}
    \hfill
    \begin{subcaptionblock}{0.49\textwidth}
        \includegraphics[width=\linewidth]{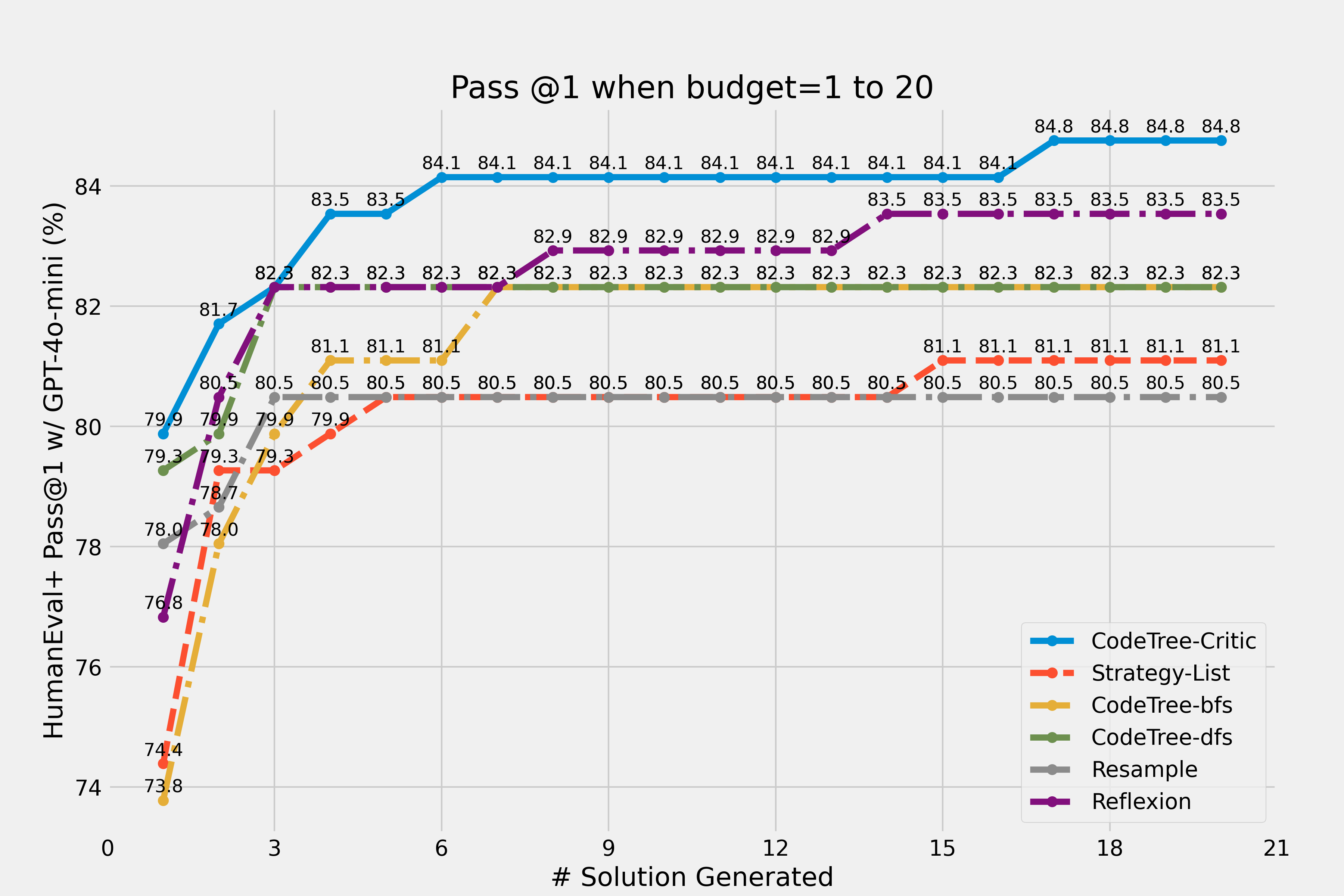}
        \caption{Performance on HumanEval+ with GPT-4o-mini}
    \end{subcaptionblock}
    \caption{Cumulative pass@1 curves while new solutions are generated within budget$=20$.}
    \label{fig:curve}
\end{figure*}
\subsection{Ablation Study}
Results in Table \ref{tab:main_res} indicate, Critic Agent plays a crucial role in CodeTree over naive search methods like BFS/DFS. We further analyzed which task in Section \ref{sec:critic} is the most crucial for Critic Agent. Specifically, we conducted the following ablation experiments: (1) w/o Solution Verification, where we excluded the verification task for any solution passing visible tests; (2) w/o Node Abort Evaluation, where we let the agents keep exploring till we reach the max depth or whenever a solution is accepted;
(3) w/o Node Scoring, where the environmental feedback is solely execution outputs, without Critic Agent's evaluation. 
\begin{table}
    \centering
    \small
    \begin{tabular}{lcc}
        \toprule
        \textbf{Model} & \multicolumn{2}{c}{\textbf{GPT-4o-mini}} \\
        \cmidrule(lr){2-3}
        \textbf{Benchmark} & \small HumanEval & \small HumanEval+ \\
        \midrule
        \textbf{CodeTree} & 94.5\% & 84.8\% \\
        w/o verification & 91.5\% & 81.7\% \\
        w/o node abort & 91.5\% & 81.1\% \\
        w/o scoring & 92.7\% & 82.1\% \\
        \bottomrule
        
    \end{tabular}
     \caption{Ablation study for different tasks of Critic Agent. We used GPT-4o-mini to evaluate corresponding settings and reported the pass@1 on the HumanEval and HumanEval+ benchmarks.}
     \label{tab:ablate}
\end{table}

The results in Table \ref{tab:ablate} show that all proposed tasks are crucial for Critic Agent to guide the tree expanding and solution search. Among these tasks, node abort and solution verification tasks are the most effective and have significant impacts on final performances. We include a qualitative study on a real example in Appendix \ref{sec:appendix}. 

\subsection{Search Efficiency}
While the performance of CodeTree is very strong with a generation budget of 20 samples, it is important to understand how our tree search strategy are more efficient than other related methods. 
Specifically, we conducted experiments when limiting the solution generation budget between 1 to 20 samples. 
In Figure \ref{fig:curve}, we plot the pass@1 curves w.r.t the number of sampled solutions: (a) shows results on CodeContests with GPT-4o; (b) shows results on HumanEval+ with GPT-4o-mini. 
In both figures, not only does our proposed CodeTree perform better than other methods, but also achieves a relatively high pass@1 even when the generation budget is small (e.g. $<9$ samples). 
On CodeContests, even when CodeTree starts with a lower performance with limited generation budgets (i.e. $<5$ samples), its performance soon improves significantly during later exploration of the tree. This observation shows that CodeTree has the potential to continue solving more problems given larger solution generation budgets.

\subsection{Results on SWEBench}
Recently, \citet{jimenez2024swebench} introduced a novel coding task for generating code patches given an input Github repository and the corresponding pull request. 
We attempted to adapt CodeTree to this benchmark by making two major changes: 
first, we replaced the problem specification with the textual description of input pull request;
secondly, we adapted the strategy generation stage as the retrieval stage where we instructed the Thinker agent to explore relevant code contexts (by code files, methods, or lines of code) from the input Github repository.  
We extended the implementation of the retrieval and code patch generation stages from \citet{xia2024agentless} and integrated our CodeTree framework for the exploration of different trajectories (from context retrieval to code patch generation and ranking). 
From Table \ref{tab:swebench}, we observed that CodeTree can lead to significant performance gains as compared to related approaches like CoT \cite{wei2022chain} and Reflexion \cite{shinn2023reflexion}. 
The results also demonstrate the versatility of our method on complex coding tasks like repo-level code generation which often requires extremely large search space to find an optimal solution. 

\begin{table}[t]
\centering 
\small
\begin{tabular}{cc}
\hline
Approach              & \% Resolved \\
\hline
\cite{xia2024agentless}             & 24.2\%       \\
\cite{xia2024agentless} + CoT       & 23.6\%       \\
\cite{xia2024agentless} + Reflexion & 25.3\%       \\
\cite{xia2024agentless} + CodeTree  & \textbf{27.6\%}      \\
\hline
\end{tabular}
\caption{Results on the SWEBench benchmark: All methods using GPT4o-mini as the base model. Compared to CoT and Reflexion, CodeTree can lead to more significant performance gain. 
}
\label{tab:swebench}
\end{table}

\section{Conclusion}
We introduce CodeTree, a new framework of agent-guided tree search for code generation tasks. 
Introducing a tree-based structure as a unified search space, CodeTree includes a Critic agent to guide the tree search and make critical decisions such as termination, expanding and scoring of tree nodes. 
CodeTree facilitates multi-agent collaboration (among Thinker, Solver, and Debugger agents) to find the correct solution within a limited solution generation budget. 
In our comprehensive experiments, CodeTree exhibits consistent and strong performance against many baselines across diverse code generation benchmarks. 

\section{Limitations}
The main limitation of this work is the requirement for LLMs with strong reasoning and instruction-following capabilities.
Empirically, we found that smaller models such as those with 8B parameters struggle at following complex instructions to play different agentic roles.
For instance, to play the role of the Critic agent, smaller models may generate unexpected output for some tasks like node scoring or self-reflection, resulting in misleading or noisy feedback to guide other LLM agents. 
In addition, incorporating agent-generated guidance will incur additional cost as it requires LLMs to extract information from long code context and generating more output tokens. 
Finally, this work focuses on the functionality correctness of code generation tasks while ignoring other qualities of generated code such as its readability or efficiency. 
To improve these qualities of output code, CodeTree can be further enhanced to incorporate more holistic LLM-generated feedback and perform diverse exploration to find optimal code solutions.

\section{Ethical Considerations}
The primary ethical concern of this paper is that directly executing the pipeline may pose a risk of information leakage or compromise the security of one’s system.
\begin{itemize}
    \item Directly passing the execution error information to commercial large language models might cause the leak of private information like the execution paths, username and package versions in the environment.
    \item Running AI-generated programs locally might put one's system and data at risk.
\end{itemize}
We will address the above issues by adding implementation of virtual environment for code execution by the time we publish our codebase.

\bibliography{custom}

\newpage
\appendix

\section{Appendix}
\label{sec:appendix}
\subsection{Qualitative Results}
\begin{figure*}[htbp]
    \centering
        \includegraphics[width=\linewidth]{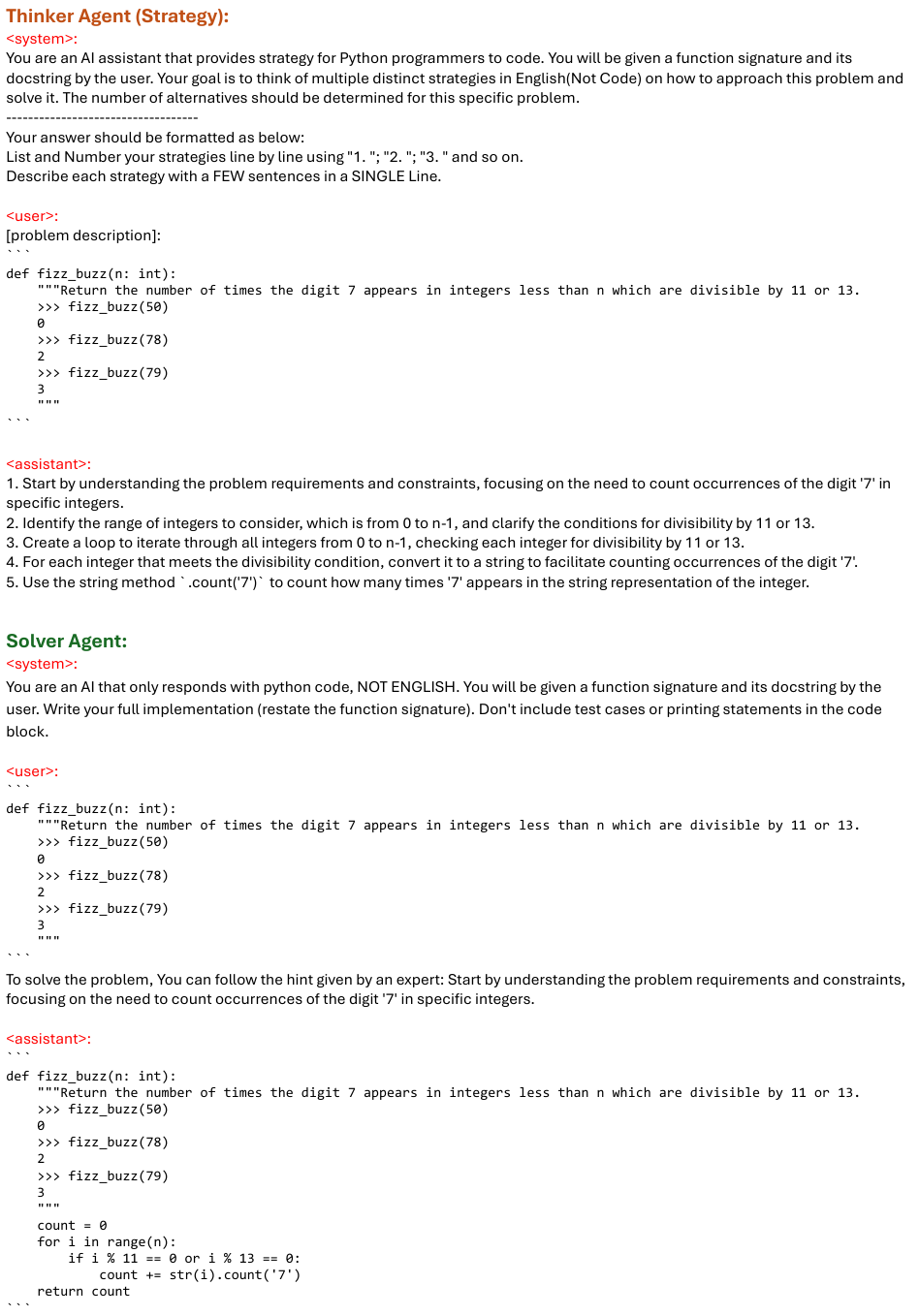}
        \caption{GPT-4o-mini as the Thinker and Solver Agents to solve HumanEval-36. The thinker agent generates 5 distinct strategies and the solver agent implements the first one. By oracle evaluation, the resulted solution can pass visible but fail on hidden test cases. }
        \label{fig:thinker-solver-prompt}
\end{figure*}

\begin{figure*}[htbp]
    \centering
        \includegraphics[width=\linewidth]{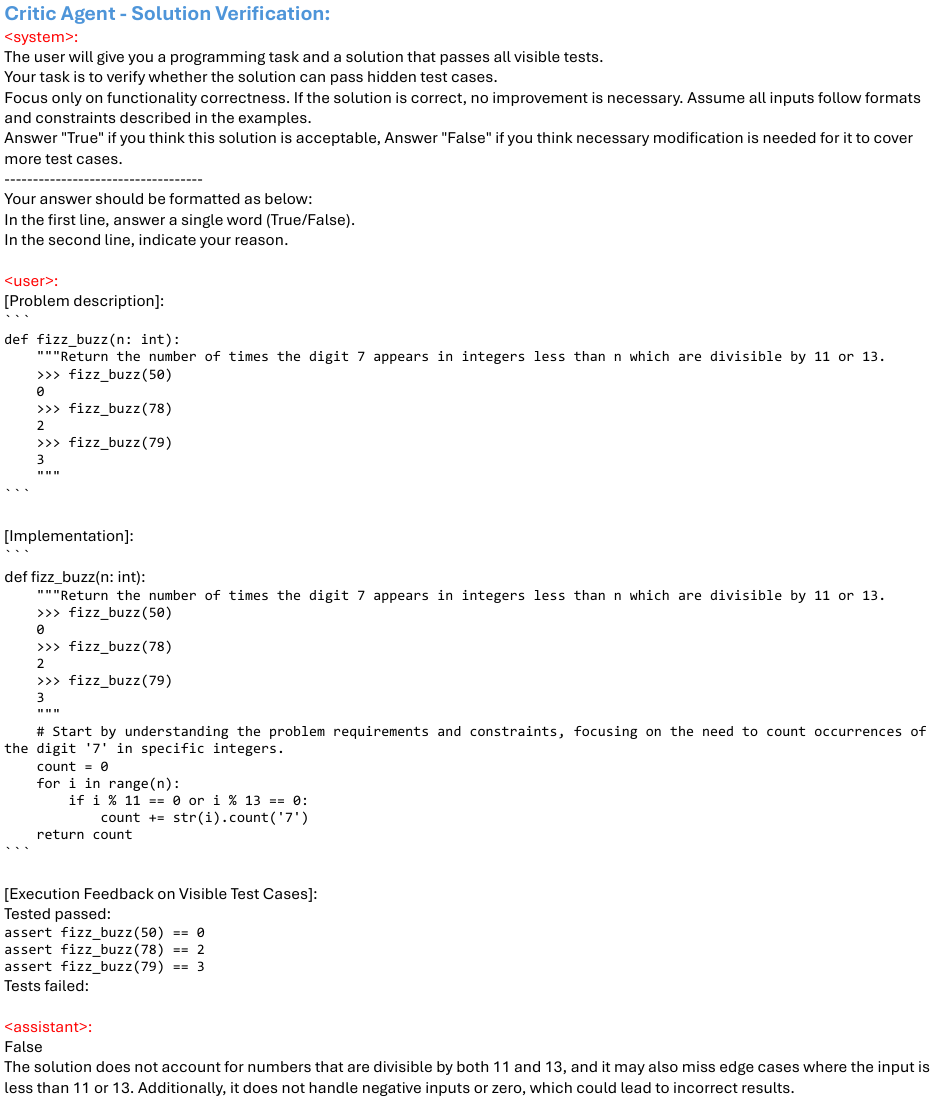}
        \caption{GPT-4o-mini as the Critic Agent to verify a solution of HumanEval-36 given by Solver Agent. It decides to reject this solution and suggest improvements.}
        \label{fig:critic-verify-prompt}
\end{figure*}

\begin{figure*}[htbp]
    \centering
        \includegraphics[width=\linewidth]{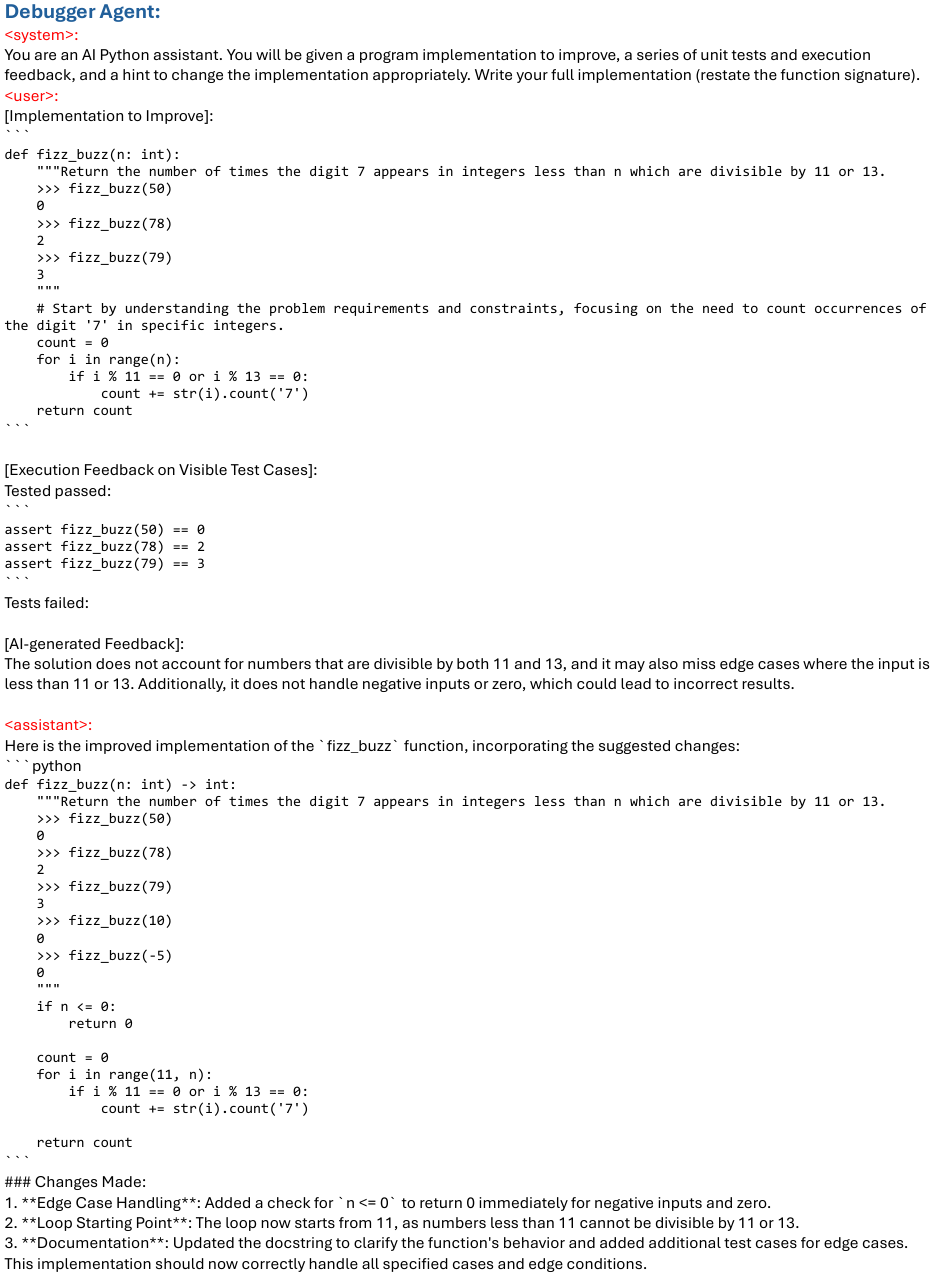}
        \caption{GPT-4o-mini as the Debugger agent to refine a solution of HumanEval-36 given by Solver Agent. It refers to Critic Agent's suggestion and correct the solution successfully.}
        \label{fig:debugger-prompt}
\end{figure*}

Here we show one real example using GPT-4o-mini to solve the problem: HumanEval-36. We keep the actual prompts and the model's responses in Figure \ref{fig:thinker-solver-prompt}, \ref{fig:critic-verify-prompt}, and \ref{fig:debugger-prompt}. We find that the initial solution $W_1$ generated, which passes all visible test cases, is rejected by the critic agent, who also suggest improvements. Debugger follows critic agent's suggestion and implement $W_2$. We evaluate $W_1$ and $W_2$ on the hidden test cases, \textbf{where $W_2$ passes and $W_1$ fails}. This indicates that the Critic Agent is making correct judgment for false positive solution(i.e., pass visible but fail on hidden test cases), and effectively guide to the correct solution. 

In Figure \ref{fig:thinker-solver-prompt}. The Thinker Agent decide to generate 5 strategies that potentially be the solutions. To start the tree search, the first strategy $S_1$ was sent to the solver agent to implement. After one solution is generated, it was sent to the Solver to be implemented. $W_1$ is immediately executed on visible test cases and observe the outputs. 

In Figure \ref{fig:critic-verify-prompt}, for a $W_1$ that passes all visible test cases, it will go to the critic agent to be further verified if $W_1$ can be applied to more general test cases. The Critic Agent gives negative judgment and suggest to improve this solution by considering more situation like zeros and negative integers. We use oracle hidden test cases to evaluate $W_1$, which is indeed incorrect, endorsing the decision made by Critic Agent. 

In Figure \ref{fig:debugger-prompt}, the Debugger Agent will treat the judgement and reason from Critic Agent as additional environmental feedback, and then implement a new solution conditioned on: 1. the problem itself, 2.the solution to improve, 3. environmental feedback. Debugger Agent refines $W_1$ to $W_2$, which is accepted by the Critic Agent. We evaluate $W_2$ on hidden test cases, it is correct.

\subsection{Other Prompts}
\begin{figure*}[htbp]
    \centering
        \includegraphics[width=\linewidth]{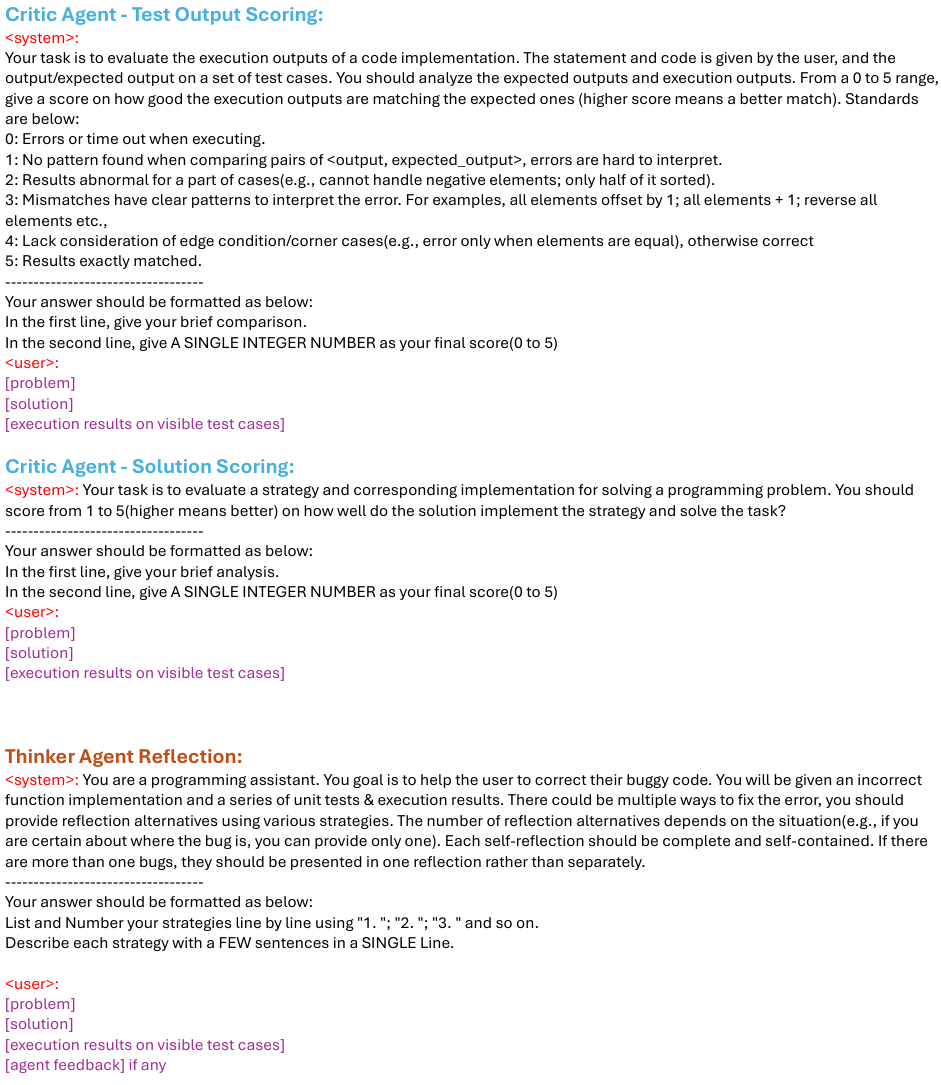}
        \caption{Prompts for Critic Agent-test cases scoring, solution scoring, as well as Thinker Agent Reflection.}
        \label{fig:other_prompt}
\end{figure*}

Critic Agent-test cases scoring, solution scoring, and Thinker Agent Reflection are not included in the above real case(HumanEval-36) solving, we presented the detailed prompts used for these agents in Figure \ref{fig:other_prompt}. 

\end{document}